Original Paper
Joe Li[1]; Peter Washington[1,*]

[1]Information and Computer Sciences, University of Hawaiʻi at Mānoa, Honolulu, HI, United States

*Corresponding author

# A Comparison of Personalized and Generalized Approaches to Emotion Recognition Using Consumer Wearable Devices: Machine Learning Study

## Abstract

**Background**: Studies have shown the potential adverse health effects, ranging from headaches to cardiovascular disease, associated with long-term negative emotions and chronic stress. Since many indicators of stress are imperceptible to observers, the early detection and intervention of stress remains a pressing medical need. Physiological signals offer a non-invasive method of monitoring emotions and are easily collected by smartwatches. Existing research primarily focuses on developing generalized machine learning-based models for emotion classification.
**Objective**: We aim to study the differences between personalized and generalized machine learning models for three-class emotion classification (neutral, stress, and amusement) using wearable biosignal data.
**Methods**: We developed a convolutional encoder for the three-class emotion classification problem using data from WESAD, a multimodal dataset with physiological signals for 15 subjects. We compared the results between a subject-exclusive generalized, subject-inclusive generalized, and personalized model.
**Results**: For the three-class classification problem, our personalized model achieved an average accuracy of 95.06% and F1-score of 91.71, our subject-inclusive generalized model achieved an average accuracy of 66.95% and F1-score of 42.50, and our subject-exclusive generalized model achieved an average accuracy of 67.65% and F1-score of 43.05.
**Conclusions**: Our results emphasize the need for increased research in personalized emotion recognition models given that they outperform generalized models in certain contexts. We also demonstrate that personalized machine learning models for emotion classification are viable and can achieve high performance.

**Keywords:** deep learning; machine learning; emotion recognition; wearable technology; affective computing; stress detection; neural network; affect detection; digital health

## Introduction

Stress and negative affect can have long-term consequences on physical and mental health, such as chronic illness, higher mortality rates, and major depression [1-3]. Therefore, the early detection and corresponding intervention of stress and negative emotions greatly reduces the risk of detrimental health conditions appearing later in life [4]. Since negative stress and affect can be difficult for humans to observe [10-12], automated emotion recognition models can play an important role in healthcare. Affective computing can also facilitate digital therapy and advance the development of assistive technologies for autism [5, 6, 27-30].

Physiological signals, including electrocardiography (ECG), electrodermal activity (EDA), and photoplethysmography (PPG), have been shown to be robust indicators of emotions [7-9]. The non-invasive nature of physiological signal measurement makes it a practical and convenient method for emotion recognition. Wearable devices such as smartwatches have become increasingly popular, and products such as Fitbit have already integrated the continuous sensing of heart rate, ECG, and EDA data into their smartwatches. The accessibility of wearable devices indicates that an emotion recognition model based on physiological data can have practical applications in healthcare.

The vast majority of research in recognizing emotions from physiological data involves machine learning models that are generalizable, which means that models were trained on one group of subjects and tested on a separate group of subjects [13, 15, 20-23, 41-46]. However, generalized models require large amounts of annotated data to train, and individuals inherently experience and react to affective stimuli differently. This creates inter-subject data variance, which degrades model accuracy [16]. Studies emphasize the need for personalized or subject-dependent models [17, 25], but it is still unclear how their performance compares against generalized models.

We present a personalized machine learning model for the three-class emotion classification problem (neutral, stress, and amusement) on the Wearable Stress and Affect Dataset (WESAD), which is the only publicly available dataset that includes both stress and emotion data [15]. To the best of our knowledge, there has been no study comparing the performance of a generalized model to a personalized model for this task. Therefore, we compare the performances of a personalized and a generalized model. We establish two generalized model baselines: one is subject-inclusive and the other subject-exclusive. Ultimately, we conclude that a personalized machine learning approach outperforms a generalized approach.

## Methods

### Overview
To classify physiological data into the neutral, stress, and amusement classes, we developed a machine learning framework and evaluated the framework using data from the Wearable Stress and Affect Dataset (WESAD). Our machine learning pipeline consists of data pre-processing, a convolutional encoder for feature extraction, and a feed-forward neural network for supervised prediction (Figure 1). Using this model architecture, we compared generalized and personalized approaches to the three-class emotion classification task (neutral, stress, and amusement).

Figure 1. Overview of our model architecture for the three-class emotion classification task.

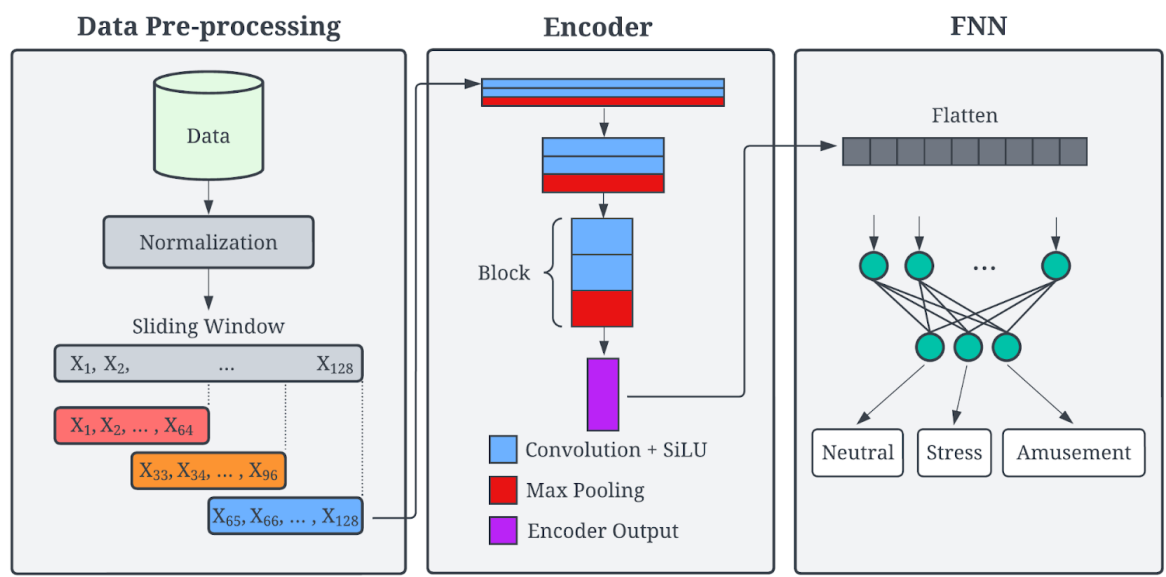

**Dataset**

We selected the Wearable Stress and Affect Dataset (WESAD) because, to the best of our knowledge, it is the only publicly available dataset that combines both stress and emotion annotations. WESAD consists of multimodal physiological data in the form of continuous time-series data for 15 subjects and corresponding annotations of four affective states: neutral, stress, amusement, and meditation. However, we only considered the neutral, stress, and amusement classes since the objective of WESAD is to provide data for the three-class classification problem, and the benchmark model in WESAD ignores the meditation state as well. Our model incorporated data from eight modalities recorded in WESAD: ECG, EDA, EMG, RESP, TEMP, and ACC (x, y, and z axes). Measurements for each of the eight modalities were sampled at 700 Hz to enforce uniformity, and data were collected for approximately 36 minutes per subject.

Figure 2. A comparison of different generalized and personalized approaches to the three-class emotion classification task. The subject-exclusive generalized model mimics generalized approaches used in other papers.

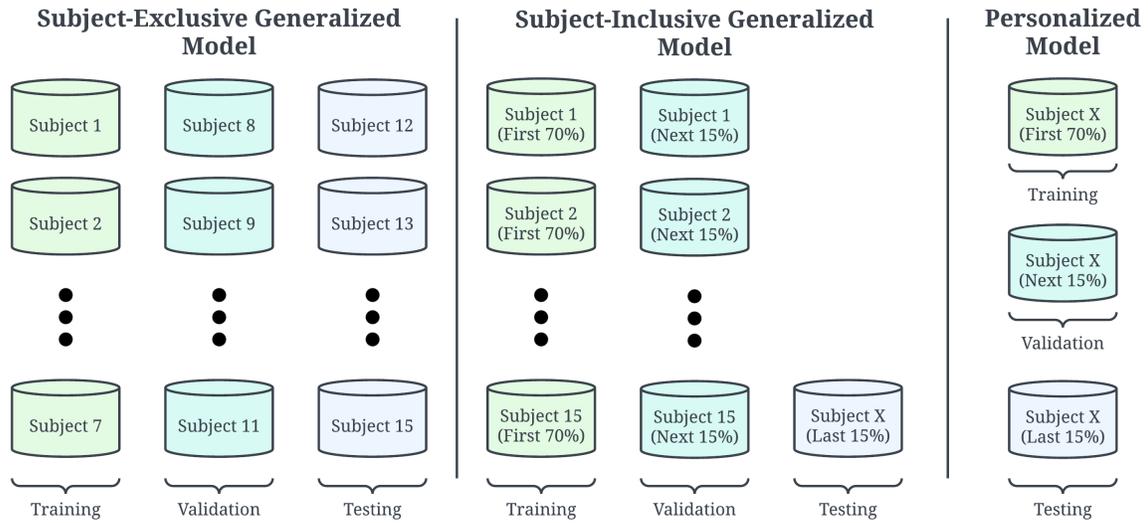

## Pre-processing and Partitioning

Each data modality was normalized with a mean of zero and a standard deviation of 1. We used a sliding window algorithm to partition each modality into intervals consisting of 64 data points, with a 50% overlap between consecutive intervals. We ensured that all 64 data points within an interval shared a common annotation, which allowed us to assign a single affective state to each interval. The process of normalization, followed by a sliding window partition, is illustrated in Figure 1. Then, these intervals were partitioned into training, validation, and testing sets.

For the personalized model, we partitioned the training, validation, and testing sets as follows. Each subject in the dataset had their own model, trained, validated, and tested independently of other subjects. For each affective state (neutral, stress, and amusement), we allocated the initial 70% of intervals with that affective state for training, the next 15% for validation, and the final 15% for testing. This guaranteed that the relative frequencies of each affective state were consistent across all three sets. Simply using the first 70% of all intervals for the training data would skew the distribution of affective states given the nature of the WESAD dataset. Furthermore, our partitioning of intervals according to sequential time order, rather than random selection, helped prevent overfitting by guaranteeing that two adjacent intervals with similar features would be in the same set. The partitioning of training, validation, and testing sets for the personalized model is shown in Figure 2.

Standard generalized models partition the training, validation, and testing sets by subject [15]. We denote these standard models as subject-exclusive generalized models, as shown in Figure 2. Through this partitioning method, it is impossible to compare the performances of generalized and personalized models since they are solving two separate tasks. Therefore, we present a modified subject-exclusive generalized model that solves the same task as the personalized model. The testing set for our subject-exclusive generalized model consisted of the last 15% of intervals for each affective state for one subject. The training set consisted of the first 70% of intervals for each affective state for all subjects except the one subject in the testing set, and the validation set consisted of the next 15% of intervals for all subjects except the one subject in the testing set. The training and testing sets for this approach contained data from mutually exclusive sets of subjects–this is where the name of the model, subject-exclusive, is derived from. Since the testing sets for the subject-exclusive generalized and personalized model are equivalent, it is possible to compare generalized and personalized approaches. This subject-exclusive generalized model served as our first generalized model baseline.

A second generalized model baseline was created, called the subject-inclusive generalized model. Like the testing sets for the subject-exclusive generalized and personalized models, the testing set for this model contained the last 15% of intervals for each affective state for a single subject. The training set consisted of the first 70% of intervals for each affective state for all subjects, and the validation set consisted of the next 15%. The set of subjects in the training and testing sets overlapped by one subject–the subject in the testing set–which is why this model is called the subject-inclusive generalized model. This is illustrated in Figure 2.

### Model Architecture

The model architecture consisted of an encoder network followed by a feed forward head, which is shown in Figure 1. Eight channels, representing the eight modalities we used from WESAD, served as input into an encoder network, which was modeled after the encoder section of U-Net [18]. The encoder network had three blocks, with each block consisting of two one-dimensional convolutional layers (kernel size of three) followed by a one-dimensional max pooling (kernel size of two). The output of each convolution operation was passed through a Sigmoid Linear Unit (SiLU) activation function. Between each block, we doubled the number of channels and added a dropout layer (15%) to prevent overfitting. The output of the encoder was flattened and passed through two fully connected layers with SiLU activation to produce a three-class probability distribution. Table 1 shows the hyperparameters that determine the model structure. These were consistent between the subject-exclusive generalized, subject-inclusive generalized, and personalized models.

Table 1. Hyperparameters relating to model structure

| Hyperparameter | Value |
| --- | --- |
|  |  |
| Encoder Depth (Number of blocks) | 3 |
| Dropout Rate | 15% |
| Number of Fully Connected Layers | 2 |
| Convolutional Kernel Size | 3 |

| Max Pooling Kernel Size | 2 |
|---|---|
| Activation Function | SiLU |

## Model Training

We trained the two generalized baseline models and the personalized model under the same hyperparameters to guarantee a fair comparison. Both models were trained with cross-entropy loss using the AdamW optimization. All models were written using PyTorch [31]. Within 1000 epochs, models with the lowest validation loss were saved for testing. A Nvidia GeForce RTX 4090 GPU was used for training. A separate personalized model was trained for each of the 15 subjects. The subject-exclusive generalized model was trained 15 times, and the subject-inclusive generalized model was trained once. For model comparison, all models were tested on each of the 15 subjects.

## Results

For the three-class emotion classification task (neutral, stress, and amusement), Figure 3 and Figure 4 illustrate the accuracy and $F_1$-score of the personalized and generalized models when tested on each of the 15 subjects. In order to guarantee a fair comparison between the models, they had the same random seeds for model initialization, and their architecture and hyperparameters were the same. The accuracy and $F_1$-score for the personalized model exceeded that of the subject-inclusive generalized model for all subjects except subject 1, and the personalized model outperformed the subject-exclusive generalized model in terms of accuracy and $F_1$-score for all subjects. Table 2 shows the average and standard deviation of the accuracies and $F_1$-scores across all subjects for the three models. We achieved an average accuracy of 95.06%, 66.95%, and 67.65% for the personalized, subject-inclusive generalized, and subject-exclusive generalized models, respectively. We also achieved an average $F_1$-score of 91.72, 42.50, and 43.05 for the personalized, subject-inclusive generalized, and subject-exclusive generalized models, respectively. Observing the error margins in Table 2, the differences in accuracy and $F_1$-score between the personalized model and both generalized models are statistically significant.

Table 2. Average accuracy and $F_1$-score of models across all subjects

| Model type | Accuracy (SD; %) | $F_1$-score (SD; %) |
|---|---|---|
|  |  |  |
| Personalized | 95.06 (9.24) | 91.72 (15.33) |
| Subject-Inclusive Generalized | 66.95 (13.76) | 42.50 (17.37) |
| Subject-Exclusive Generalized | 67.65 (13.48) | 43.05 (17.20) |

Figure 3. A comparison of model accuracy between the personalized and generalized models.

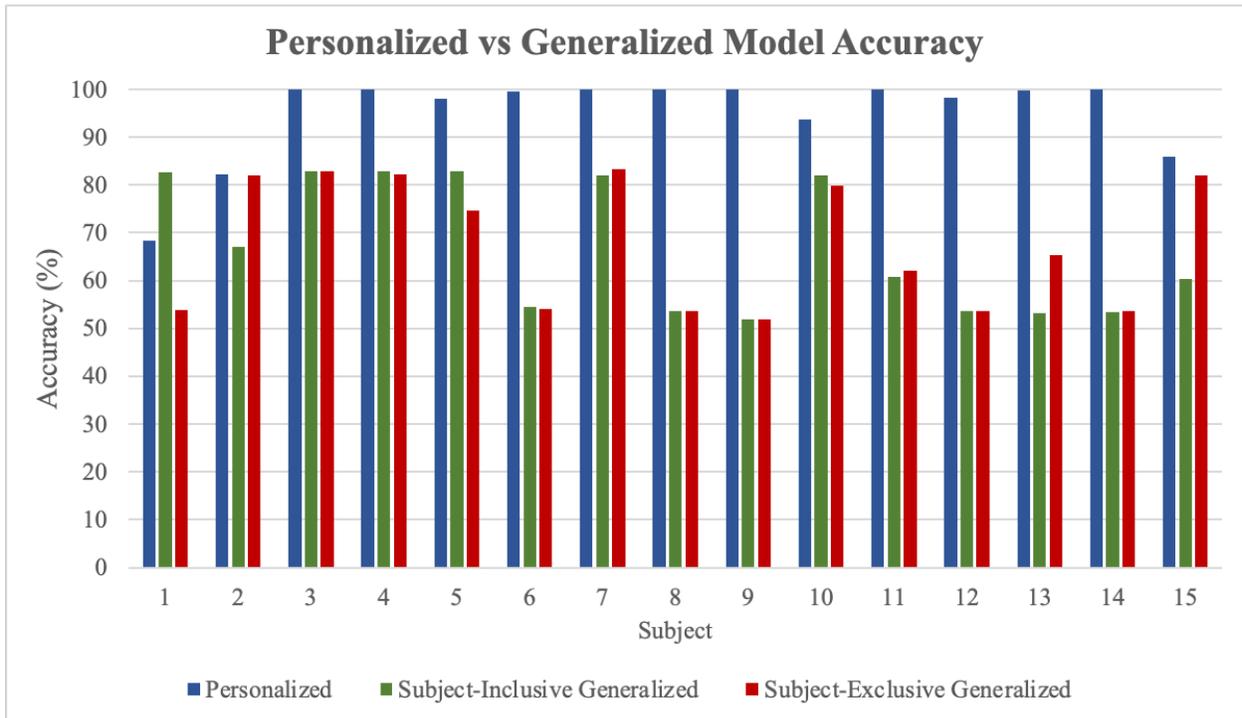

Figure 4. A comparison of F$_1$-score between the personalized and generalized models.

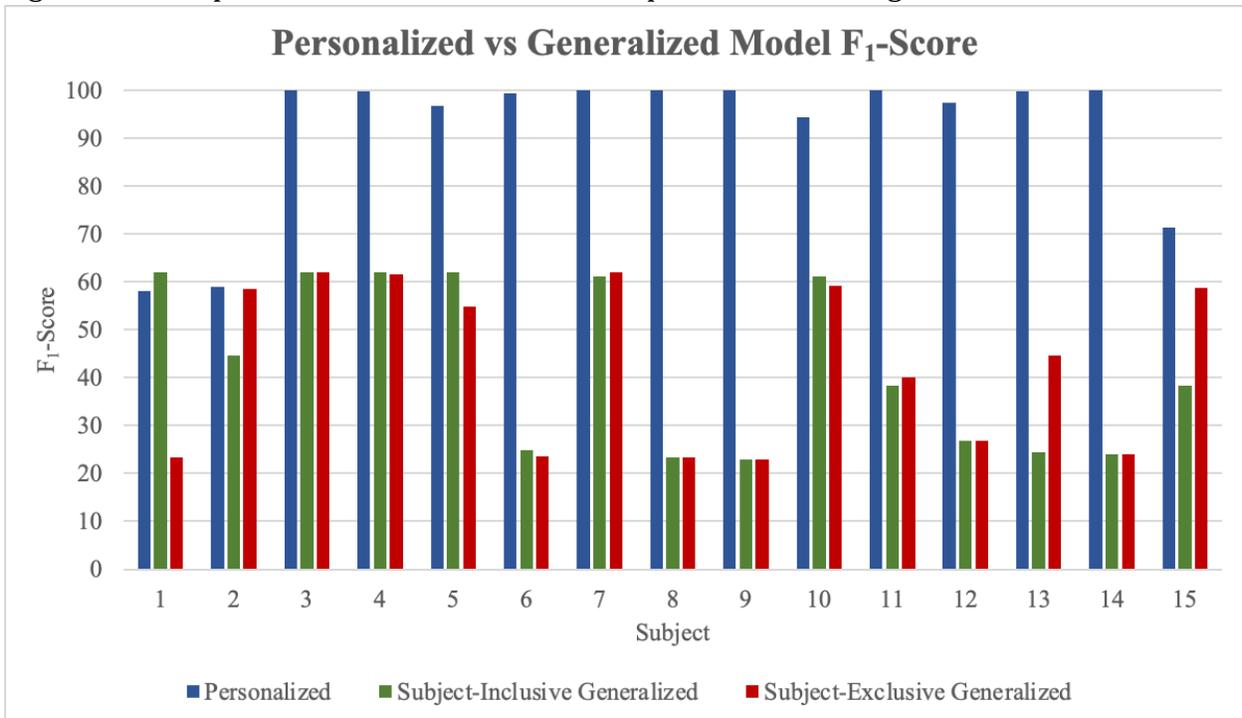

# Discussion

## Principal Findings

Using a machine learning approach with a convolutional encoder, we demonstrated that a personalized model outperforms a generalized model in both the accuracy and $F_1$-score metrics for the three-class emotion classification task. By establishing two generalized model baselines through the subject-inclusive and subject-exclusive models, we created an alternative approach to the standard generalization technique of separating the training and testing sets by subject, and as a result, we were able to compare personalized and generalized approaches. Our personalized model achieved an accuracy of 95.06% and $F_1$-score of 91.72, while our subject-inclusive generalized model achieved an accuracy of 66.95% and $F_1$-score of 42.50 and our subject-exclusive generalized model achieved an accuracy of 67.65 and $F_1$-score of 43.05.

Our work indicates that personalized models for emotion recognition should be further explored in the realm of healthcare. Machine learning methods for emotion classification are clearly viable and can achieve high accuracies, as shown by our personalized model. Furthermore, given that numerous wearable technologies collect physiological signals, data acquisition is both straightforward and non-invasive. Combined with the popularity of consumer wearable technology, it is feasible to scale emotion recognition systems. This can ultimately play a major role in the early detection of stress and negative emotions, thus serving as a preventative measure for serious health problems ranging from major depression to heart attacks.

## Comparison with Prior Work

### Generalized Models

The vast majority of studies developed generalized approaches to the emotion classification task. Schmidt et al [15], the pioneer of WESAD, created several feature extraction models and achieved accuracies up to 80% for the three-class classification task. Huynh et al [23] developed a deep neural network, trained on WESAD wrist signals, to outperform past approaches by 8.22%. Albaladejo-González et al [35] achieved a $F_1$-score of 88.89 using an unsupervised local outlier factor model and 99.03 using a supervised multi-layer perceptron. Additionally, they analyzed the transfer learning capabilities of different models between the WESAD and SWELL-KW [36] datasets. Ghosh et al [37] achieved 94.8% accuracy using WESAD chest data by encoding time-series data into Gramian Angular Field images and employing deep learning techniques. Bajpai et al [38] investigated the k-nearest neighbor algorithm to explore the tradeoff between performance and total number of nearest neighbors using WESAD. Through federated learning, Almadhor et al [39] achieved 86.82% accuracy on data in WESAD using a deep neural network. Behinaein et al [40] developed a novel transformer approach and achieved state-of-the-art performance using only one modality from WESAD.

### Personalized Models

Sah and Ghasemzadeh [25] developed a generalized approach using a convolutional neural network using one modality from WESAD. For the three-class classification problem, they achieved an average accuracy of 92.85%. They used the leave-one-subject-out (LOSO) analysis to highlight the need of personalization. Indikawati and Winiarti [33] directly developed a personalized approach for the four-class classification problem in WESAD (neutral, stress, amusement, meditation). Using different feature extraction machine learning models, they achieved accuracies ranging from 88-99% for the 15 subjects. Liu et al [34] developed a federated learning approach using data from WESAD with the goal of preserving user privacy. In doing so, they developed a personalized model as a baseline, which achieved an average accuracy of 90.2%. Despite studies in both the generalized and personalized approaches, there exists no direct comparison between a generalized and personalized model. Our paper therefore bridges this gap and concretely demonstrates how personalization outperforms generalization.

### Limitations and Future Work

One limitation of our work is that we only evaluated data for 15 subjects from a single dataset (WESAD). It is valuable to reproduce results on additional physiological signal datasets for emotion analysis, such as DEAP [14] and CLAS [19]. Furthermore, data from WESAD were collected under controlled laboratory environments, which may not generalize to the real world. Therefore, analyzing emotions in a real-world context through datasets such as K-EmoCon [24], which contain physiological data collected during naturalistic conversations, may be useful. Emotions in the K-EmoCon dataset were categorized into 18 different classes, so exploring this dataset could also help us better assess the benefits of personalization for a broader range of emotions. A major goal of this approach is to provide support for personalized digital interventions for neuropsychiatry which could benefit a variety of applications such as video-based digital therapeutics for children with autism [47-54].

For personalized models, individuals must annotate their own data, which can be time-consuming and expensive. To reduce the burden of manual labeling, self-supervised learning approaches can be explored, where a model first pre-trains on a large set of unlabeled data. Then, the pre-trained model can be fine-tuned to a small amount of labeled data provided by the user. This self-supervised learning framework follows that of natural language processing models such as BERT [27]. Studies have begun to explore the tradeoff between the amount of labels and model accuracy. For example, Khan and Sarkar [32] compared a supervised algorithm to a semi-supervised generative adversarial network trained on partially-labeled data from WESAD, and they demonstrated that partial labeling can still maintain accurate results.

### Conflicts of Interest

All authors declare no competing interests.

## Abbreviations

ACC: acceleration
BERT: Bidirectional Encoder Representations from Transformers
CLAS: Cognitive Load, Affect, and Stress
DEAP: Database for Emotion Analysis using Physiological Signals
ECG: electrocardiogram
EDA: electrodermal activity
EMG: electromyogram
FNN: feed-forward neural network
PPG: photoplethysmogram
RESP: respiration
SiLU: Sigmoid Linear Unit
TEMP: temperature
WESAD: Wearable Stress and Affect Dataset